%% file: main.tex
\documentclass{article}

\usepackage[preprint,nonatbib]{neurips_2023}

\usepackage[utf8]{inputenc} 
\usepackage[T1]{fontenc}    
\usepackage{hyperref}       
\usepackage{url}            
\usepackage{booktabs}       
\usepackage{amsfonts}       
\usepackage{nicefrac}       
\usepackage{microtype}      
\usepackage{xcolor}         

\input{macros}

\usepackage{graphicx} 
\usepackage{amsmath,amssymb,amsthm}
\usepackage{booktabs}
\usepackage{hyperref}
\usepackage{cleveref}
\usepackage{color}
\usepackage{bbold}
\usepackage{graphicx}
\usepackage{algorithm}
\usepackage{array}
\definecolor{colorMomentum}{RGB}{239, 134, 91}
\definecolor{colorPolarization}{RGB}{78, 115, 174}
\definecolor{colorMedian}{RGB}{82, 111, 170}
\definecolor{colorCI}{RGB}{169, 184, 212}
\definecolor{colorDemoReturn}{RGB}{170, 50, 3}
\definecolor{colorBckReturn}{RGB}{35, 107, 174}

\title{Discovering Individual Rewards in Collective Behavior through Inverse Multi-Agent Reinforcement Learning}

\author{%
Daniel Wälchli$^{1,2}$, Pascal Weber$^{1,2}$, Petros Koumoutsakos$^{2}$\thanks{Corresponding Author: {petros@seas.harvard.edu}} \\[1em]
  ${}^1$ Computational Science and Engineering Laboratory, ETH Zürich, Switzerland \\
  ${}^2$ John A. Paulson School of Engineering and Applied Sciences, Harvard University, USA
}

\begin{document}

\maketitle

\begin{abstract}
The discovery of individual objectives in collective behavior of complex dynamical systems such as fish schools and bacteria colonies is a long-standing challenge. 
Inverse reinforcement learning is a potent approach for addressing this challenge but its applicability to dynamical systems, involving continuous state-action spaces and multiple interacting agents, has been limited. 
In this study, we tackle this challenge by introducing an off-policy inverse multi-agent reinforcement learning algorithm (IMARL). 
Our approach combines the ReF-ER techniques with guided cost learning. 
By leveraging demonstrations, our algorithm automatically uncovers the reward function and learns an effective policy for the agents.
Through extensive experimentation, we demonstrate that the proposed policy captures the behavior observed in the provided data, 
and achieves promising  results across problem domains including single agent models in the OpenAI gym and multi-agent models of schooling behavior.
The present study shows that the proposed IMARL algorithm is a significant step towards understanding collective  dynamics from the perspective of its constituents, and showcases its value as a  tool for studying complex physical systems exhibiting collective behaviour.

\end{abstract}

\section{Introduction}
\label{sec:intro}
Collective behavior is a hallmark of social and natural systems including crowds~\cite{Helbing2005}, markets~\cite{Vriend1995}, schooling behavior of animals~\cite{Aoki1982}, and even artificial multi-agent systems~\cite{multiagentsystems}. In these systems the coordinated actions or behaviors exhibited by the group of individuals, is the result of interactions between individuals that are in turn influenced by the actions and behaviors of others within the group.

Inferring the  individual objectives based on the emerging phenomenon remains  an elusive task. Inverse reinforcement learning (IRL) automates  the process of finding a reward function based on observations of behaviour. Arguably the  reward  is also the most transferable object in a decision-making process~\cite{Russell1998}. Here we propose  IRL for  collective behavior by extending prior work on forward Multi-Agent Reinforcement Learning (MARL). More specifically we hybridize the Remember and Forget Experience Replay (ReF-ER) for cooperative MARL~\cite{ReFER-MARL2022} with Guided Cost Learning (GCL)~\cite{GCL2016}. While both algorithms have demonstrated great success in several applications~\cite{novati2021a,bae2022a,Hu2023}, their combination, to the best of our knowledge, is a novel contribution. This hybridization allows us to solve the IRL problem in a continuous state- and action space with multiple agents for computationally challenging tasks. We apply this algorithm first to a single agent setting in three OpenAI gym MuJoCo environments~\cite{brockman2016openai,mujoco}. The extension to multiple agents is then shown by an application to a particle model of schooling behavior. By discovering  a reward function for the agents, we can successfully reproduce their collective behavior.

The paper is organized as follows: In~\cref{sec:mairl} we introduce MARL and GCL as well as algorithmic implementation details. In~\cref{sec:results}, we present the validation and the results for the collective behavior of particle systems and present the conclusion in~\cref{sec:conclusion}.

\subsection{Related Work}

Reinforcement learning has successfully tackled challenges in various domains, including games~\cite{Mnih2015, Silver2017}, control~\cite{Lillicrap2015}, scientific machine learning~\cite{brunton2020a}, and natural language modeling~\cite{Ouyang2022}. However, when studying animal and human behavior, the reward function is often unknown~\cite{Russell1998}. IRL was introduced to address this issue, but it revealed that multiple reward functions can explain the same observed behavior~\cite{Ng2000}. Nevertheless, IRL has been used effectively to learn controllers for acrobatic helicopter maneuvers from observational data~\cite{Abbeel2004, Abbeel2006}.
Later works have formulated the IRL problem from a stochastic perspective~\cite{Ramachandran2007,Ziebart2008}. By leveraging the maximum entropy assumption and neural networks to model the reward function, more complex applications became feasible~\cite{Levine2012, Wulfmeier2015, Ho2016}. Guided cost learning (GCL)\cite{GCL2016}, which has shown success in robotics\cite{Ravichandar2020, Hu2023}, forms the foundation of the present work. Despite these successes, research on multi-agent IRL with continuous action spaces is limited~\cite{Lantao2019, Wonseok2020, liu2022distributed}. Previous works in the context of animal behavior~\cite{Sosic2016, Yamaguchi2018,Xin2021,ashwood2022dynamic} rely on the discretization of the action space, on on-policy methods, or the linearization of dynamics, which are not suitable for computationally expensive applications. In this study, we narrow this gap by combining GCL with ReF-ER\cite{Novati2019}, a well-established algorithm for scientific machine learning that has demonstrated success in challenging flow problems~\cite{gunnarson2021a,amoudruz2021a,novati2021a,bae2022a}.

\section{Methods}\label{sec:mairl}

To introduce Inverse Multi-Agent Reinforcement Learning  (IMARL) we  first  describe the forward problem. The forward MARL can be  described as a  Multi-Agent Markov Decision Process (MAMDP)~\cite{MARL}. Here,  all agents share the same state space, action space, and reward. In this setting, the MAMDP is described by a tuple $\mathcal{M}=(\stS, \acS, \re, D, N)$ consisting of the state-space $\stS$, the action-space $\acS$, the reward function $\re$, the transition map $D$, and the number of agents $N$. The agents' individual states, actions, and rewards at time steps $t=1,\dots,T$ are denoted by $\st_t^{(i)}\in\stS$, $\ac_t^{(i)}\in\acS$, and $\re_t^{(i)}\in\mathbb{R}$ for $i=1,\dots,N$, and we denote the respective collective states/actions/rewards for all agents by $\vstate_t,\vaction_t$, and $\vreward_t$. Following these definitions, we assume that the transition map takes the form $\vreward_t,\vstate_{t+1}=D(\vstate_t,\vaction_t)$. Furthermore, we define the stochastic policy ${\pi}(\ac|\st)$, which is a probability distribution over the action space given the state for an individual agent. This follows the decentralized execution paradigm, in which agents act solely based on their local state~\cite{Gupta2017}. For each agent, we can define the state-value as
\begin{equation}
{V}^{\pi}(\st^{(i)})=\mathbb{E}_{\pi}\left[\sum\limits_{t=0}^{\infty}\gamma^tr_t^{(i)}|\st^{(i)}_0=\st^{(i)}\right]\COMMA
\end{equation}
where $\gamma\in [0,1)$ is the discount factor. The goal of collaborative MARL with experience sharing is to find the optimal policy $\pi^\star$ that maximizes the average of the state values for all agents 
\begin{equation}
{\pi}^\star=\arg\max\limits_\pi\frac{1}{N}\sum\limits_{i=1}^N V^{\pi}(\sti)\COMMA\quad \forall s^{(0)},\dots,s^{(N)}\in{\cal S}\PERIOD
\end{equation}
The policy $\pi$  is approximated as $\pi_{\omega}$ by a neural network  having weights $\omega$. We identify the 
optimal policy using  ReF-ER MARL~\cite{ReFER-MARL2022}, the multi-agent extension of V-RACER with ReF-ER~\cite{Novati2019}. ReF-ER MARL is an actor-critic off-policy MARL algorithm. It avoids outliers in the replay memory using the pointwise ratio between the online and offline policies and penalizes the policy update using the Kullback-Leibler divergence. For further details, we refer the reader to the original publications.

In IMARL the reward function $\re$ is unknown and needs to be ascertained based on a set of observed trajectories $\DATA=\{\tau_k\}_{k=1}^K=\{\langle(\st_0,\ac_0),(\st_1,\ac_1),\dots,(\st_{T_k},\ac_{T_k})\rangle_k\}_{k=1}^{K}$. For sake of brevity we have not included the agent-index. We approximate the reward function with a neural network $\re_{\vartheta}: \stS \times \acS \rightarrow \mathbb{R}$. In order to learn the unknown parameters $\vartheta$, we employ GCL~\cite{GCL2016}. Under the maximum entropy assumption~\cite{Ziebart2008,Wulfmeier2015}, trajectories with higher returns are exponentially more preferred
\begin{equation}
    p(\tau\GIVEN\vartheta) = \frac{1}{Z_\vartheta} e^{\mathcal{R}_\vartheta(\tau)}\COMMA
\end{equation}
where $Z_\vartheta$ is the partition function and the return $\mathcal{R}_\vartheta$ of a trajectory $\tau$ is defined as the sum of the rewards $\mathcal{R}_\vartheta(\tau) = \sum_{j=0}^{T} \re_{\vartheta,j}$. The goal is to maximize the likelihood of the observed data
\begin{equation}
    p(\DATA\GIVEN\vartheta)=\prod_{k=1}^K p(\tau_k\GIVEN\vartheta) = Z_\vartheta^{-K}\prod_{k=1}^K e^{\mathcal{R}_\vartheta(\tau_k)}\PERIOD
\end{equation}
Equivalently, we can maximize the log-likelihood of the demonstrations
\begin{equation}
\label{eq:maxentobj}
\begin{split}
\mathcal{L}(\vartheta,\DATA)
&=  \sum_{k=1}^K\mathcal{R}_\vartheta(\tau_k)-K \log Z_{\vartheta}\PERIOD
\end{split}
\end{equation}
While the first term can be readily computed, the partition function is approximated as described in the following section. The goal of IRL is to find the parameters $\vartheta^\star$ of the optimal reward function $\re_\star$
\begin{equation}
    \vartheta^\star=\arg\max\limits_\vartheta \mathcal{L}(\vartheta,\DATA)\PERIOD
\end{equation}In order to better understand the objective we compute the derivative
\begin{equation}
\label{eq:maxentobj-grad}
\begin{split}
\frac{\partial \mathcal{L}(\vartheta,\DATA)}{\partial \vartheta}
&=  \sum_{k=1}^K\frac{\partial \mathcal{R}_\vartheta(\tau_k)}{\partial\vartheta}-K \frac{1}{Z_{\vartheta}}\frac{\partial Z_{\vartheta}}{\partial\vartheta}\\
&=  \sum_{k=1}^K\frac{\partial \mathcal{R}_\vartheta(\tau_k)}{\partial\vartheta}-K \frac{1}{Z_{\vartheta}}\int_{\Omega_\tau}\frac{\partial \mathcal{R}_\vartheta(\tau)}{\partial\vartheta}e^{\mathcal{R}_\vartheta(\tau)}\mathrm{d}\tau \\
&=  \sum_{k=1}^K\frac{\partial \mathcal{R}_\vartheta(\tau_k)}{\partial\vartheta}-K \mathbb{E}_{\tau\sim p(\cdot\GIVEN\vartheta)}\left[\frac{\partial \mathcal{R}_\vartheta(\tau)}{\partial\vartheta}\right]\PERIOD
\end{split}
\end{equation}
This gradient is used to update the weights of the reward using gradient ascent. While the first term maximizes the return on the demonstration trajectories, the second term acts as a regularizer which minimizes the mean return on the space of trajectories~\cite{Ramachandran2007,Ziebart2008}.

\begin{figure}
    \centering
    \includegraphics[trim=20 175 20 175, clip,width=\linewidth]{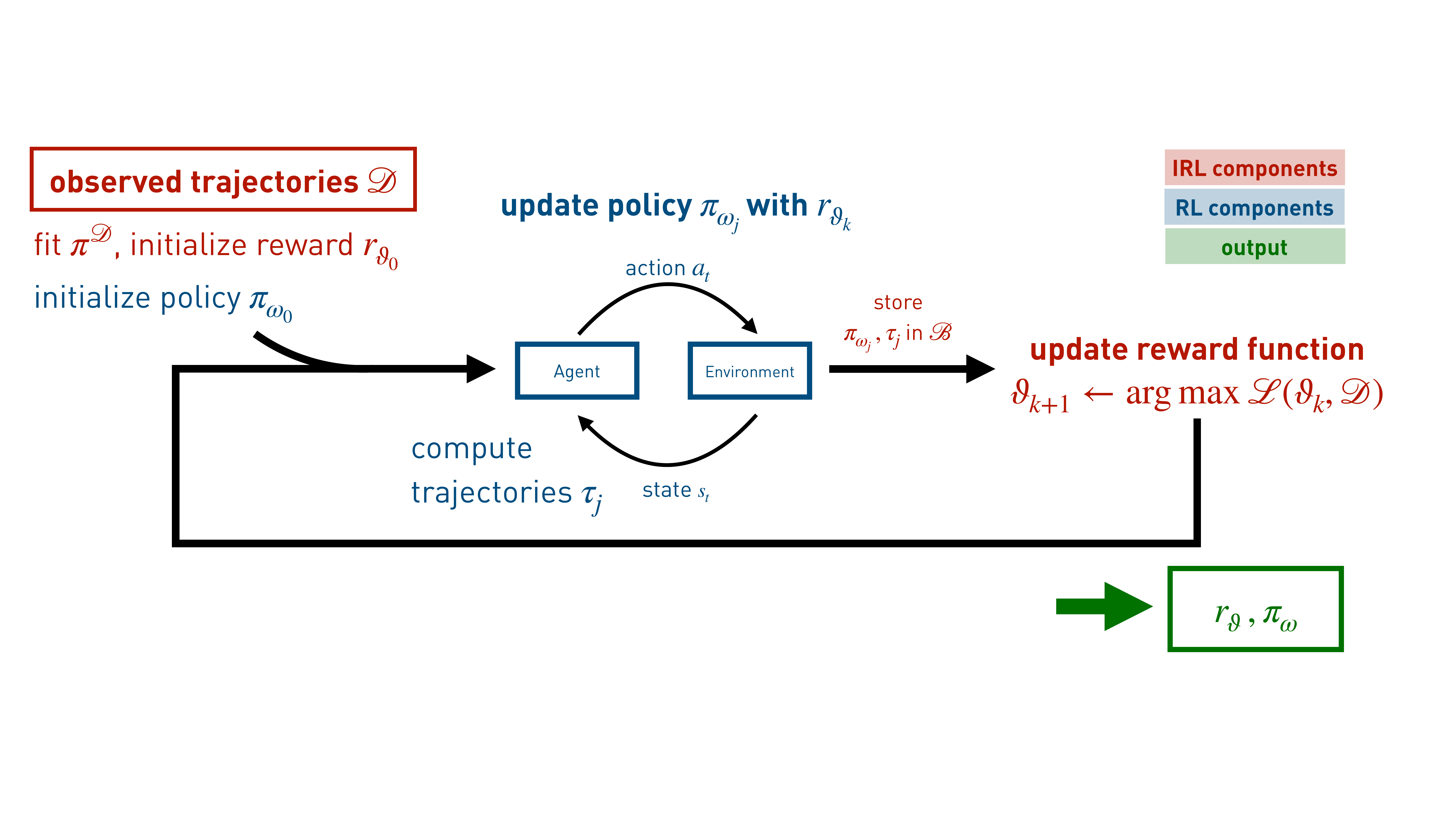}
    \caption{Schematic of the IMARL algorithm. Input are the observed trajectories $\DATA$, and the output is the learned reward function $\re_\vartheta$ plus the corresponding optimal policy $\pi_\omega$ of the agents.}
    \label{fig:ma-irl}
\end{figure}

\subsection{Approximation}
The partition function is approximated using Monte Carlo integration
\begin{equation}
\label{eq:normZ}
Z_\vartheta=\mathbb{E}_{\tau}\left[e^{\mathcal{R}_\vartheta(\tau)}\right]\approx \frac{|\Omega_{\cal T}|}{M}\sum_{m=1}^M e^{\mathcal{R}_\vartheta(\tau_m)}\COMMA\quad\tau_m\sim\mathcal{U}(\Omega_\mathcal{T})\COMMA
\end{equation}
where $\Omega_\mathcal{T}$ represents the domain of possible trajectories. Plugging this into~\cref{eq:maxentobj} gives an estimator for the objective
\begin{equation}
\label{eq:maxentobj-approx}
\begin{split}
\mathcal{L}(\vartheta,\DATA)\approx \sum_{k=1}^K\RF(\tau_k)-K \log \frac{|\Omega_{\cal T}|}{M}\sum_{m=1}^M e^{\RF(\tau_m)}\COMMA\quad\tau_m\sim\mathcal{U}(\Omega_\mathcal{T})\PERIOD
\end{split}
\end{equation}
Since the reward is expressed by  a neural network, its gradient is  readily computed through  backpropagation. To learn both, the non-linear policy of the agents $\pi_\omega$ and the reward function $\re_\vartheta$, we iterate between solving the forward problem and optimizing the reward as illustrated in~\cref{fig:ma-irl}. This allows combining trajectories collected during the \textit{forward} MARL stored in a background batch $\mathcal{B}$ and the demonstration trajectories $\mathcal{D}$, which helps estimating the partition function~\cite{GCL2016}. Since the trajectories collected in this manner follow a non-uniform distribution, they are inconsistent for the Monte-Carlo estimator~\cref{eq:maxentobj-approx}. In order to construct a consistent estimator an importance weight $\omega_m$ is computed with an inverse-linear pooling function~\cite{koliander2014}
\begin{equation}\label{eq:importanceWeight}
    \omega_m=\frac{1}{\frac{1}{M}\sum_{m'=1}^Mp_{m'}(\tau_m)}\COMMA \quad \text{with}\quad p_m(\tau)=p(s_0)\prod_{t=0}^{T-1} D(s_{t+1}\GIVEN a_t,s_t)\pi_m(a_t|s_t)\COMMA
\end{equation}
where $\pi_m$ denotes the policy used to compute trajectory $\tau_m$. Note that the computational cost of the evaluation of the importance weights grows quadratically with the number of samples, since each trajectory must be evaluated with all policies in~\cref{eq:importanceWeight}. The resulting Monte Carlo estimate then reads
\begin{equation}
\label{eq:maxentobj-iw}
\begin{split}
\mathcal{L}(\vartheta,\DATA)\approx \sum_{k=1}^K\RF(\tau_k)-K \log \frac{|\Omega_{\cal T}|}{M}\sum_{m=1}^M \omega_m e^{\RF(\tau_m)}\COMMA\quad\tau_m\sim p(\mathcal{D},\mathcal{B})\COMMA
\end{split}
\end{equation}
where $p(\mathcal{D},\mathcal{B})$ denotes the distribution of the demonstrations and background trajectories. When computing the stochastic gradient for the update, one uses
\begin{equation}
\label{eq:maxentobj-grad-approx}
\begin{split}
\frac{\partial\mathcal{L}(\vartheta,{\cal D})}{\partial \vartheta}
=  \sum_{i=1}^{M_1}\Bigg[1-&\frac{K}{Z_\theta} \frac{|\Omega_{\cal T}|}{M_1+M_2} \omega_ie^{\mathcal{R}_{\vartheta}(\tau_i)}\Bigg]\, \frac{\partial \mathcal{R}_{\vartheta}(\tau_i)}{\partial\vartheta}\\ 
&- \frac{K}{Z_\theta} \frac{|\Omega_{\cal T}|}{M_1+M_2}\sum_{j=1}^{M_2} \omega_je^{\mathcal{R}_{\vartheta}(\tau_j)}\, \frac{\partial \mathcal{R}_{\vartheta}(\tau_j)}{\partial\vartheta}\COMMA
\end{split}
\end{equation}
where $\tau_i\sim p(\mathcal{D})$ and $\tau_j\sim p(\mathcal{B})$ and $M_1$ and $M_2$ denote the mini-batch sizes for the demonstration and the background trajectories. Furthermore we note, that the partition function $Z_\vartheta$ is estimated using~\cref{eq:normZ} and the unknown $|\Omega_\tau|$ cancels. In order to determine the probabilities of the demonstration trajectories according to~\cref{eq:importanceWeight}, we fit a linear controller $\pi_\DATA(\ac|\st)$ with Gaussian error terms. In the multi-agent setting, the trajectories of the individual agents are processed sequentially. 

\subsection{Implementation}

\begin{table}[!b]
	\centering
     \caption{Hyper-parameter of the IMARL algorithm. The left columns lists the parameter related to reward function learning, and in the right columns the parameter related to policy learning are shown.}\label{tab:irl-hp}
	\begin{tabular}{m{4.5cm}m{1.6cm}|m{3.6cm}m{1.6cm}}
		\toprule \multicolumn{2}{c}{Reward Network} &  \multicolumn{2}{c}{Policy Network}
		\\ \midrule
  	Hidden Layers & 2  & Hidden Layers & 2 \\
        Width & 64  & Width & 128 \\
        Activation functions & Soft ReLu & Activation functions & Soft ReLu \\
	    Background batch size & 512 & Replay memory size &  262144 \\
		Experiences per update & 10000 & Experiences per update & 1 \\
        Optimizer & Adam & Optimizer & Adam \\
        Learning rate & 1e-4  & Learning rate & 1e-4  \\
        Mini-batch size background  & 16 & Mini-batch of experiences & 128\\
        Mini-batch size demonstration  & 16 & & \\
  \bottomrule
	\end{tabular}
\end{table}
For the \textit{forward} reinforcement learning, we uses a single neural network to approximate both, the policy and the value function. The reward function is approximated using a second feed-forward neural network. We use a shifted sigmoid function at the output of the network, such that the rewards of the agents are bound $\re_\vartheta: \stS \times \acS \rightarrow (-0.5,0.5)$. We found that constraining the reward function to a bounded domain improves the convergence rate of the reward function and the policy. During training, we update the reward function less frequently than the policy network (the update ratio is lower than $1:1000$), so the policy can adapt to the changing objective function during training. When sampling the environment, the reward function is evaluated on the state and action, and the resulting triplet is stored in the replay memory. The same approach is followed when updating the policy: to ensure that the experiences in the mini-batch are consistent with the current reward function, the states and actions are forwarded through the reward network before the stochastic gradient step of the policy. For the updates of the neural networks, we use Adam~\cite{adam}. The computational cost of the importance weight calculation is limited by fixing the size of the background batch $\BCK$. When the capacity of $\BCK$ is reached, the algorithm overwrites the oldest trajectories. We apply default values from \cite{Novati2019} for policy learning. For reward learning, we conduct a hyperparameter search to determine the optimal values of the mini-batch size of the background and demonstration trajectories, the update frequency of the reward network, and the width of the reward network. We follow the best practices for choosing hyperparameters reported in \cref{tab:irl-hp}. The search required 144 runs for a maximum duration of 24hrs on an Intel\textregistered Xeon\textregistered E5-2690 v3 @ 2.60GHz processor (12 cores, 64GB RAM). The IMARL algorithm is integrated into a distributed computing framework,  Korali~\cite{korali}, publicly available under \url{https://anonymous.4open.science/r/korali-E79F/}. This framework was also used to create the observations to validate the OpenAI gym MuJoCo tasks. The scripts to generate the observational data for the swarming experiment are available under \url{https://anonymous.4open.science/r/swarm-8CB1/}.

\section{Results}
\label{sec:results}

\subsection{Validation}~\label{sec:validation}
\begin{figure}[!b]
    \centering
    \includegraphics[width=1.\linewidth]{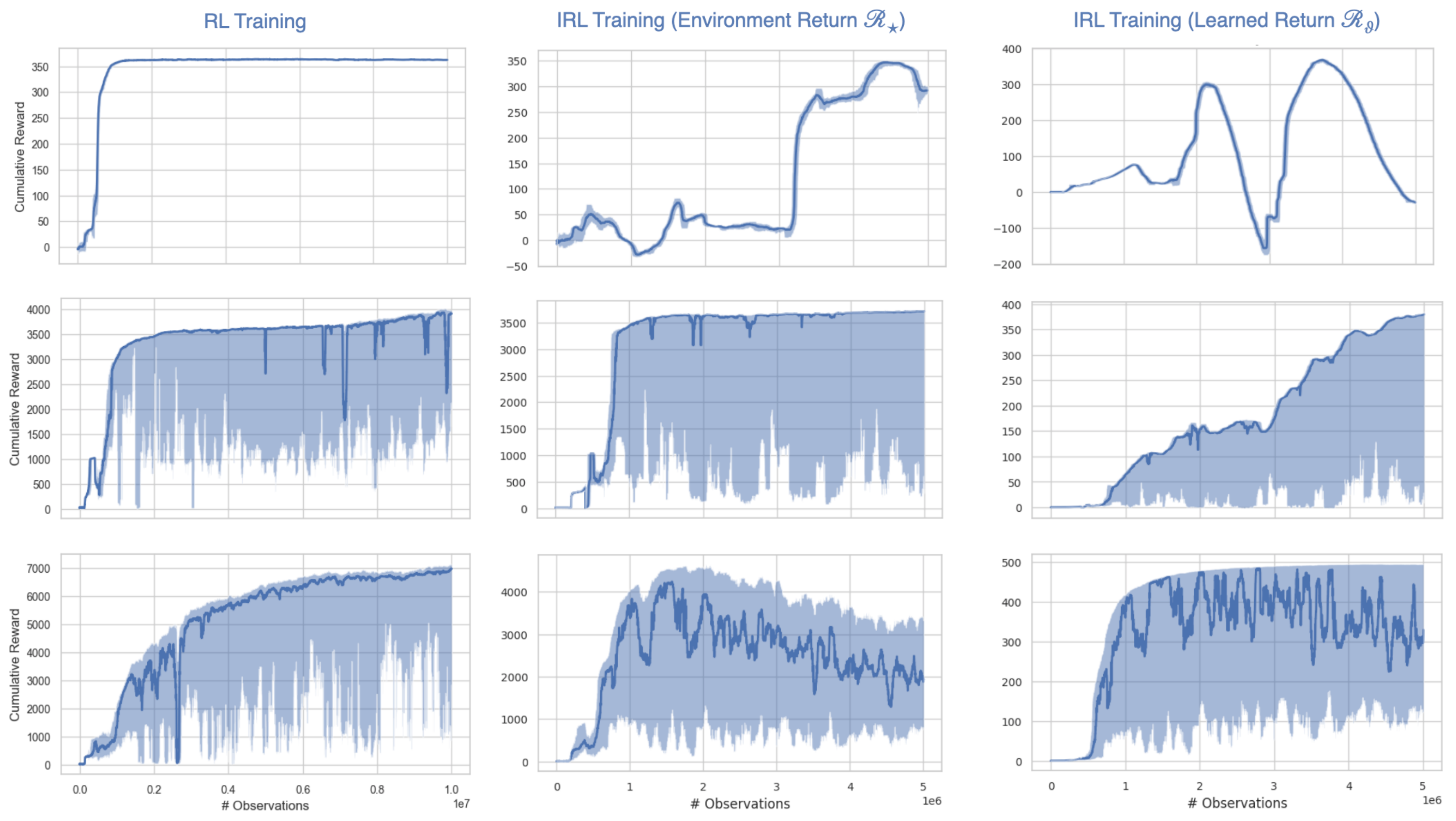}
    \caption{From top to bottom: \textit{Swimmer-v4}, \textit{Hopper-v4}, and \textit{Walker2d-v4}. From left to right: RL returns in the  environment. Environment returns recovered via IRL. Returns calculated from the learned reward function $\mathcal{R}_\vartheta$. The running median indicated with (\textcolor{colorMedian}{\rule[0.5ex]{1.em}{1.5pt}}), the shading shows the running 80\% confidence interval of the returns (\textcolor{colorCI}{\rule[0.ex]{1.em}{5pt}}) calculated from the last 100 episodes.}
    \label{fig:validation}
\end{figure}
We validate the implementation in the single agent setting on three OpenAI gym MuJoCo tasks, namely in the \textit{Swimmer-v4}, the \textit{Hopper-v4}, and the \textit{Walker2d-v4} environment. MuJoCo is a physics engine for robotic tasks where the state of the agent is a combination of coordinates and velocities of the joints, and the actions are the torques that will be applied. The reward is calculated from the maximal distance covered within 1000 steps and a penalization for strong actuation \cite{brockman2016openai, mujoco}.

We first solve the forward RL problem and learn a policy with V-RACER using ten million interactions with the environment. With the learned policy, we generate the demonstration data set $\mathcal{D}$ consisting of 100 trajectories for each environment. We then recover the reward function and a policy from the demonstration data using IRL with five million interactions with the environment. To verify the procedure,  we compare the environment returns obtained with RL and IRL in~\cref{fig:validation}. We find that in the \textit{Swimmer-v4} and the \textit{Hopper-v4} environment, we reach 95\% of the maximal return obtained with RL. In the \textit{Walker2d-v4} environment we recover 65\% of the return. The \textit{Walker2d-v4} environment has a larger state ($s\in\mathbb{R}^{17}$) and action space ($a\in\mathbb{R}^6$) compared to the \textit{Swimmer-v4} ($s\in\mathbb{R}^{8}, a\in\mathbb{R}^2$), \textit{Hopper-v4}  ($s\in\mathbb{R}^{11}, a\in\mathbb{R}^3$) and thus is a harder problem to solve. Interestingly, the returns obtained from the learned reward function $\re_\vartheta$ in the \textit{Swimmer-v4} environment show non-monotonic behavior while the maximal return is recovered. This observation can be attributed to the presence of the regularization term in \cref{eq:maxentobj-grad}, respectively in \cref{eq:maxentobj-grad-approx}. In the  \textit{Walker2d-v4} environment, we find a reward function and policy that achieves the maximal possible cumulative reward of 500 (the return has an upper bound of 500 due to the sigmoid activation function at the output of the reward function approximator and the fixed episode length 1000).

\subsection{Collective behavior}~\label{sec:application}
\begin{figure}[!b]
    \centering
    \includegraphics[width=1.\linewidth]{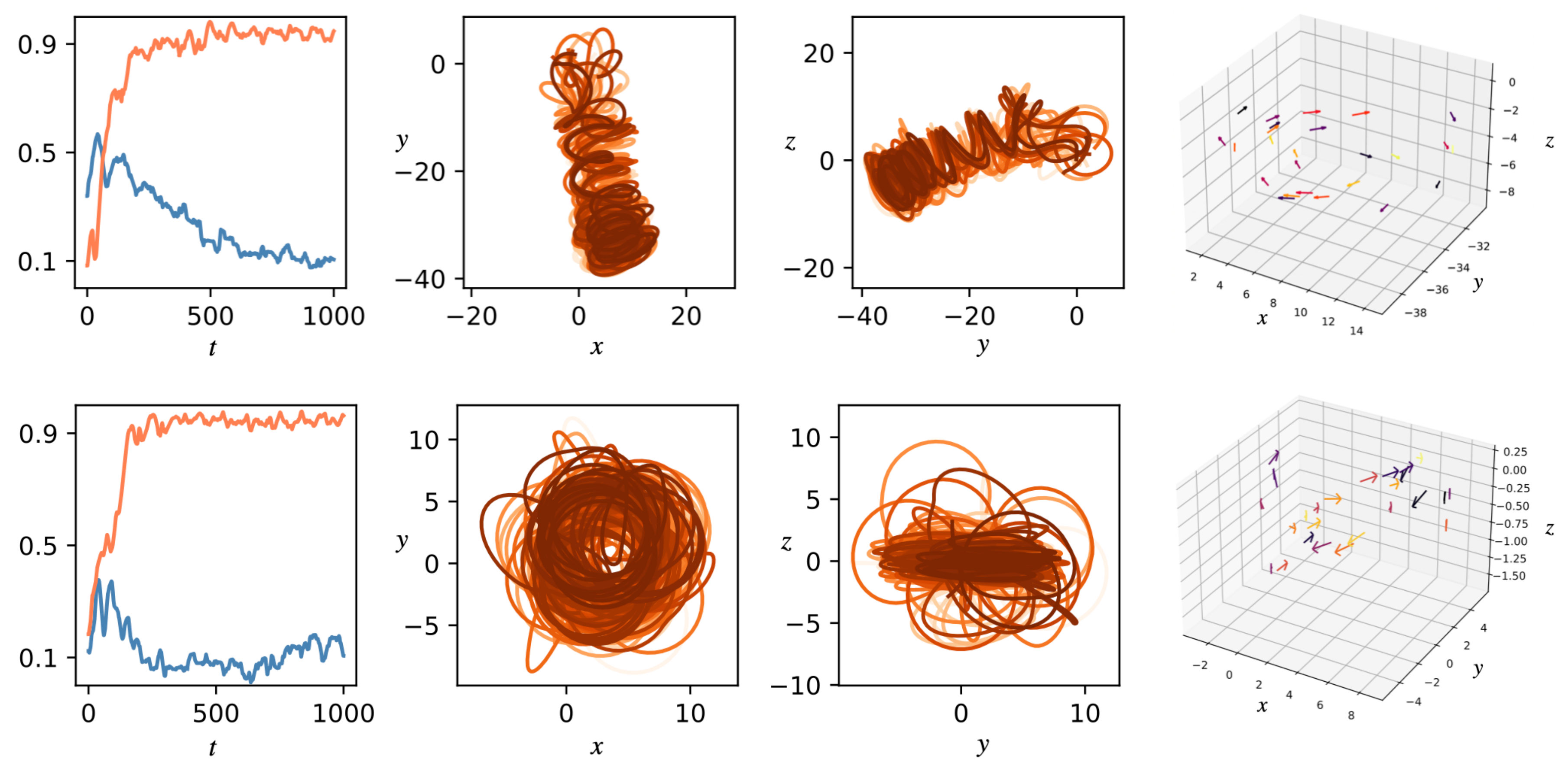}
    \caption{Two trajectories from the demonstration data $\mathcal{D}$. From left to right: Evolution rotation  (\textcolor{colorMomentum}{\rule[0.5ex]{1.em}{1.5pt}}) and polarization (\textcolor{colorPolarization}{\rule[0.5ex]{1.em}{1.5pt}}) for $t\in[0,1000)$. Trajectories of the swimmers projected onto the x-y plane. Trajectories of the swimmers projected onto the y-z plane. Location and swimming direction of the swimmers at t=1000. }
    \label{fig:demo-traj}
\end{figure}
\begin{figure}
    \centering
    \includegraphics[width=0.6\linewidth]{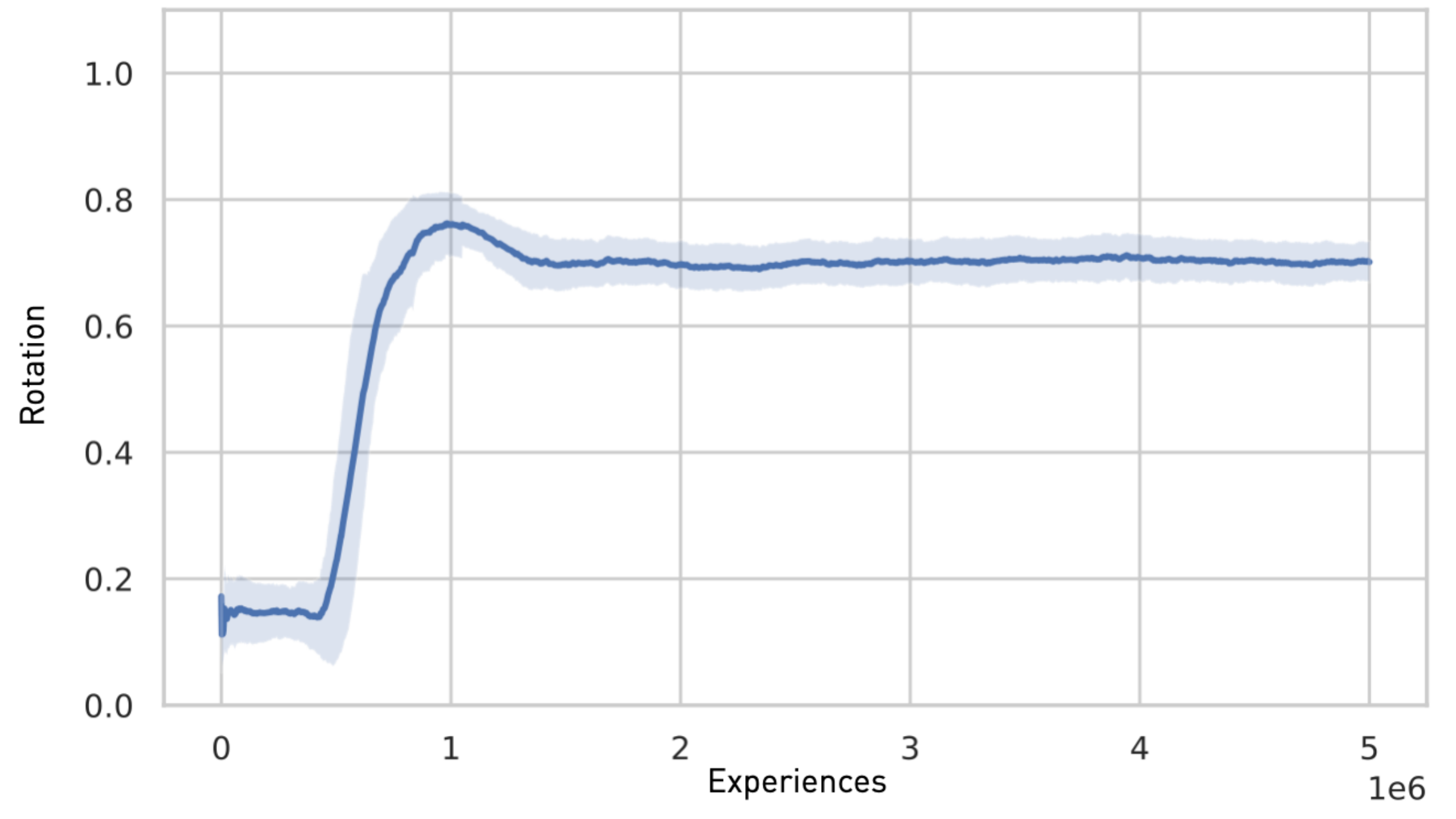}
    \includegraphics[width=0.34\linewidth]{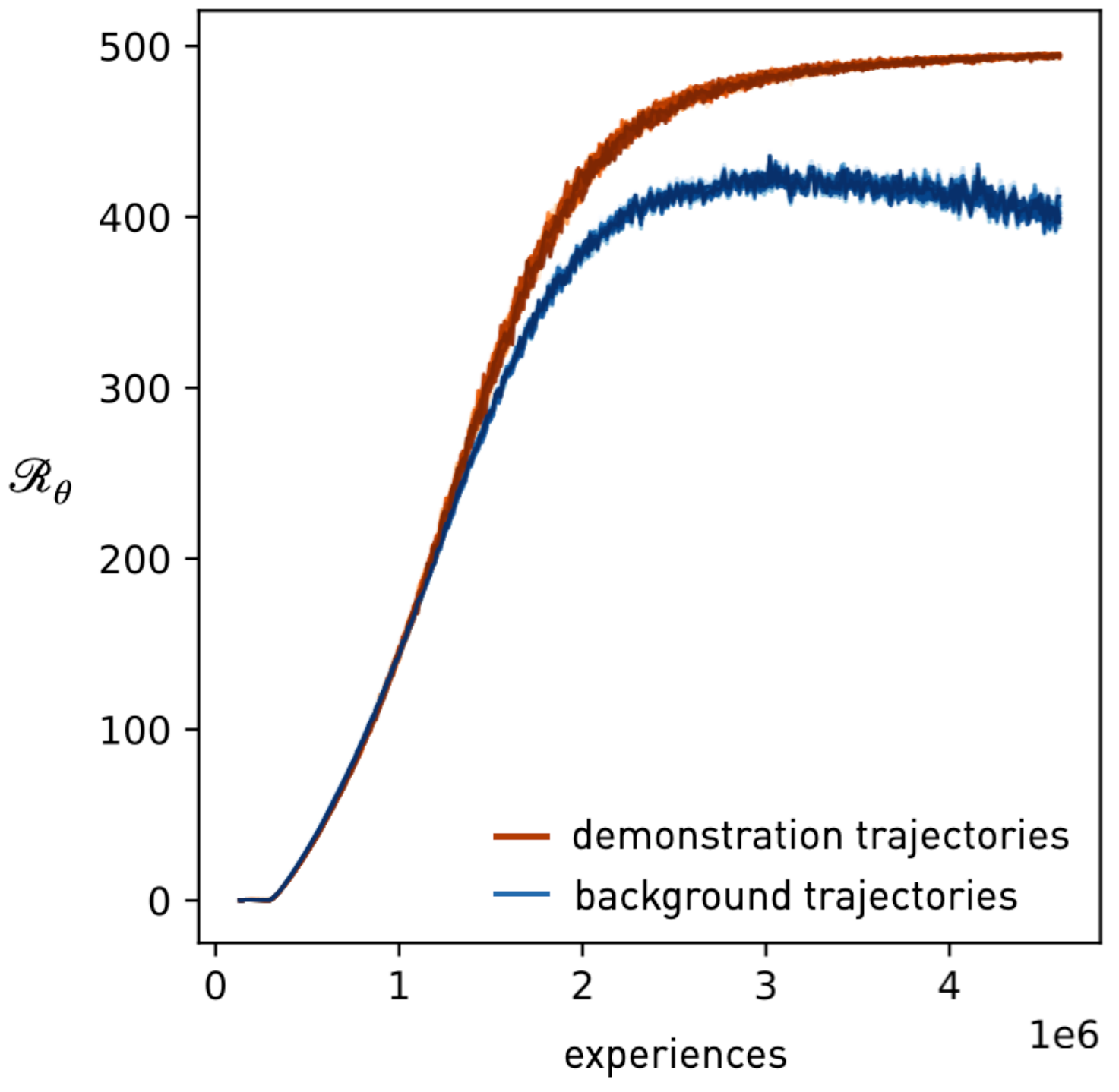}
    \caption{Illustration of the training process. The left figure shows the time-averaged rotation of the swarm observed during training. The running median indicated with (\textcolor{colorMedian}{\rule[0.5ex]{1.em}{1.5pt}}), the shading shows the running 80\% confidence interval of the rotation (\textcolor{colorCI}{\rule[0.ex]{1.em}{5pt}}) calculated from the last 100 episodes. The right figure displays the average returns of the (\textcolor{colorDemoReturn}{\rule[0.5ex]{1.em}{1.5pt}}) demonstration and (\textcolor{colorBckReturn}{\rule[0.5ex]{1.em}{1.5pt}}) background trajectories in the mini-batch during the reward function update. The different colour grades indicate the returns of the 25 agents.}
    \label{fig:training}
\end{figure}
We model the collective behavior of fish schooling using a well-established model~\cite{Aoki1982,Gautrais2008}. In the following, we will briefly describe and refer to the original publications for details.

The position and velocity of swimmer $i=1,\dots,N$ at timestep $t$ are given by $\boldsymbol{x}^{(t)}_i$ and $\boldsymbol{v}_i^{(t)}$. We denote the normalized direction of motion as $\hat{\boldsymbol{v}}_i^{(t)}$ and assume that all swimmers move at constant velocity $v=\|\boldsymbol{v}_i\|$. For the initial positioning of the swimmers, we sample from a random uniform distribution over an n-sphere. The dynamics are modeled in three steps: First, the wished direction ${\hat \vel_i^{\star}}$ is computed by defining three spherical zones: the zone of repulsion \textsc{zor}, the zone of orientation \textsc{zoo}, and the zone of attraction \textsc{zoa}. If two swimmers enter the zone of repulsion, the wished direction is chosen such that the swimmers move away from each other,
\begin{equation}
    { \vel_i^{\star}}=-\frac{\sum_j\disp_{ij}^{(t)}}{\|\sum_j\disp_{ij}^{(t)}\|}\quad\forall j\ne i\text{ with }\|\disp_{ij}^{(t)}\|<r_r\,,
\end{equation}
where $\disp_{ij}^{(t)}=\loc_j^{(t)}-\loc_i^{(t)}$ denotes the displacement vector. If there is no swimmer within the zone of repulsion, the swimmers mutually align and attract one another according to
\begin{equation}
\begin{split}
    { \vel_i^{\star}} &= \frac{\sum_{k}\vel_k^{(t)}}{\|\sum_{k}\vel_k^{(t)}\|}  + \frac{\sum_{\ell}\disp_{i\ell}^{(t)}}{\|\sum_\ell \disp_{i\ell}^{(t)}\|}\quad\forall k\text{ with }r_r < \|\disp_{ik}^{(t)}\|<r_o\;\text{and}\; \forall \ell \text{ with } r_o < \|\disp_{i\ell}^{(t)}\|<r_a\PERIOD
\end{split}
\end{equation}
We finish the computation of the wished direction by normalizing $\vel_i^{\star}$.
\begin{equation}
    {\hat \vel_i^{\star}}=\frac{\vel_i^{\star}}{\|\vel_i^{\star}\|}\,.
\end{equation}
To account for the stochastic nature of the decision-making of swimmers, we further add normally distributed noise to the process. This is imperative for collective patterns like schooling~\cite{NIWA199647}. We compute the swimming direction $\vel_i^{(t)}$ by rotating the current direction towards the wished direction. The location of the swimmers is updated with an explicit forward Euler timestep
\begin{equation}\label{eq:eul_timestep}
    \loc^{(t+\Delta t)}_i=\loc^{(t)}_i+\Delta t\;\vel_i^{(t)}\PERIOD
\end{equation}
By fine-tuning the radii, the model replicates three behaviours, swarming, schooling, and milling. Using the following two order parameters, we can quantitatively classify the behaviour: The rotation
\begin{equation}\label{eq:momentum}
    R = \frac{1}{N}\left\|\sum^{N}_{i=1}\frac{(\boldsymbol{x}_i-\boldsymbol{\mu})}{\|(\boldsymbol{x}_i-\boldsymbol{\mu})\|}\times \frac{\boldsymbol{v}_i}{v}\right\|\in[0,1]\,,\quad \text{where}\quad \boldsymbol{\mu}=\frac{1}{N}\sum^{N}_{i = 1}\boldsymbol{x}_i\COMMA
\end{equation}
and the polarization
\begin{equation}\label{eq:polarization}
    P = \frac{1}{N}\frac{1}{v}\left\|\sum^{N}_{i=1}\boldsymbol{v}_i\right\|\in[0,1]\PERIOD
\end{equation}
For swarming, both the polarization and the rotation are low. For schooling, the swimmers have a high polarization and low rotation. Milling is characterized by low polarization and high rotation.
\begin{figure}
    \centering
    \includegraphics[width=1.\linewidth]{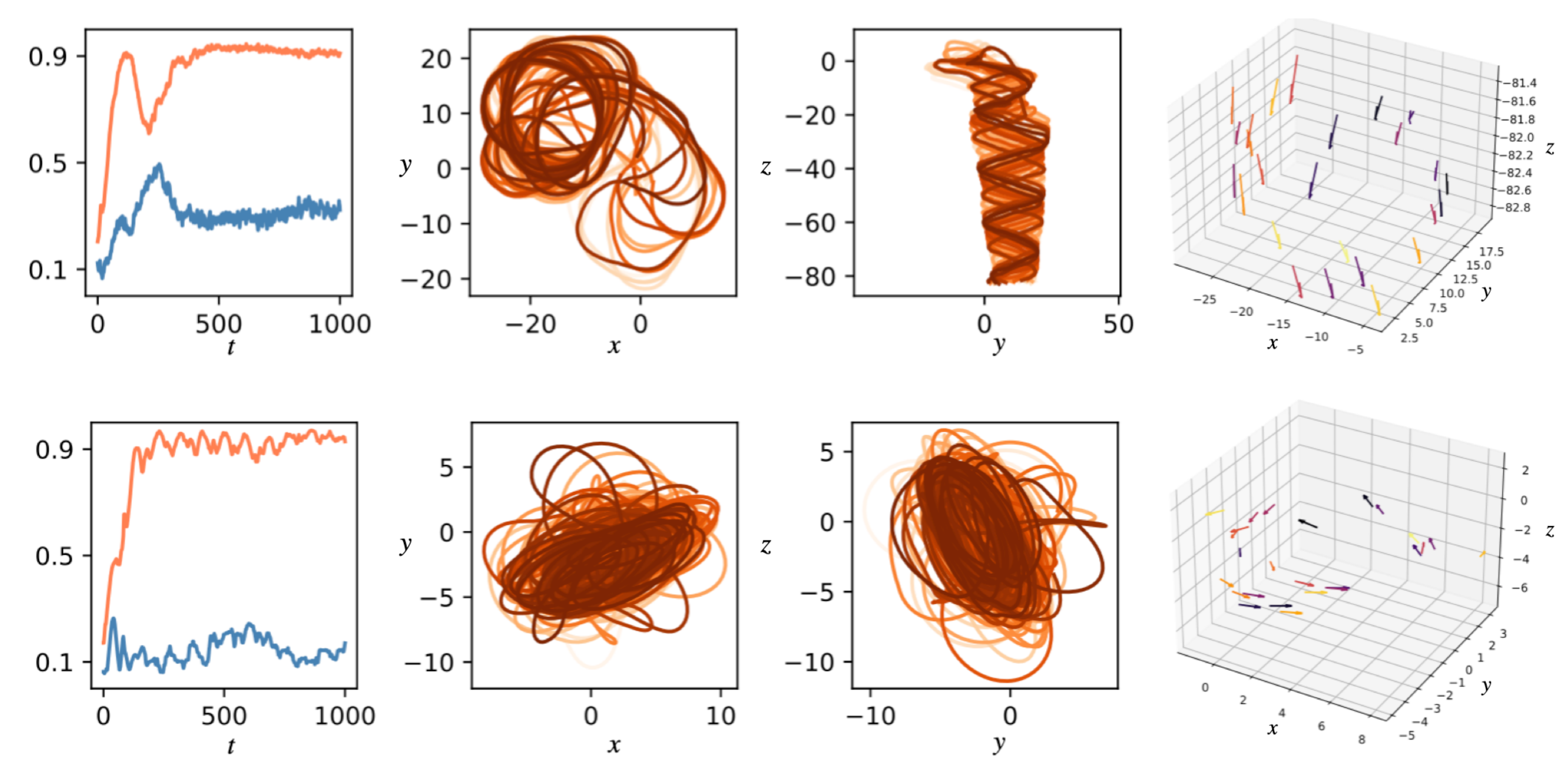}
    \caption{Two trajectories sampled from the policy $\pi_{\omega}$ learned via IRL. From left to right: Evolution of rotation (\textcolor{colorMomentum}{\rule[0.5ex]{1.em}{1.5pt}}) and polarization (\textcolor{colorPolarization}{\rule[0.5ex]{1.em}{1.5pt}}) for $t\in[0,1000)$. Trajectories of the swimmers projected onto the x-y plane. Trajectories of the swimmers projected onto the y-z plane. Location and swimming direction of the swimmers at t=1000.}
    \label{fig:irl-traj}
\end{figure}
\begin{figure}[!b]
    \centering
    \includegraphics[width=1.0\linewidth]{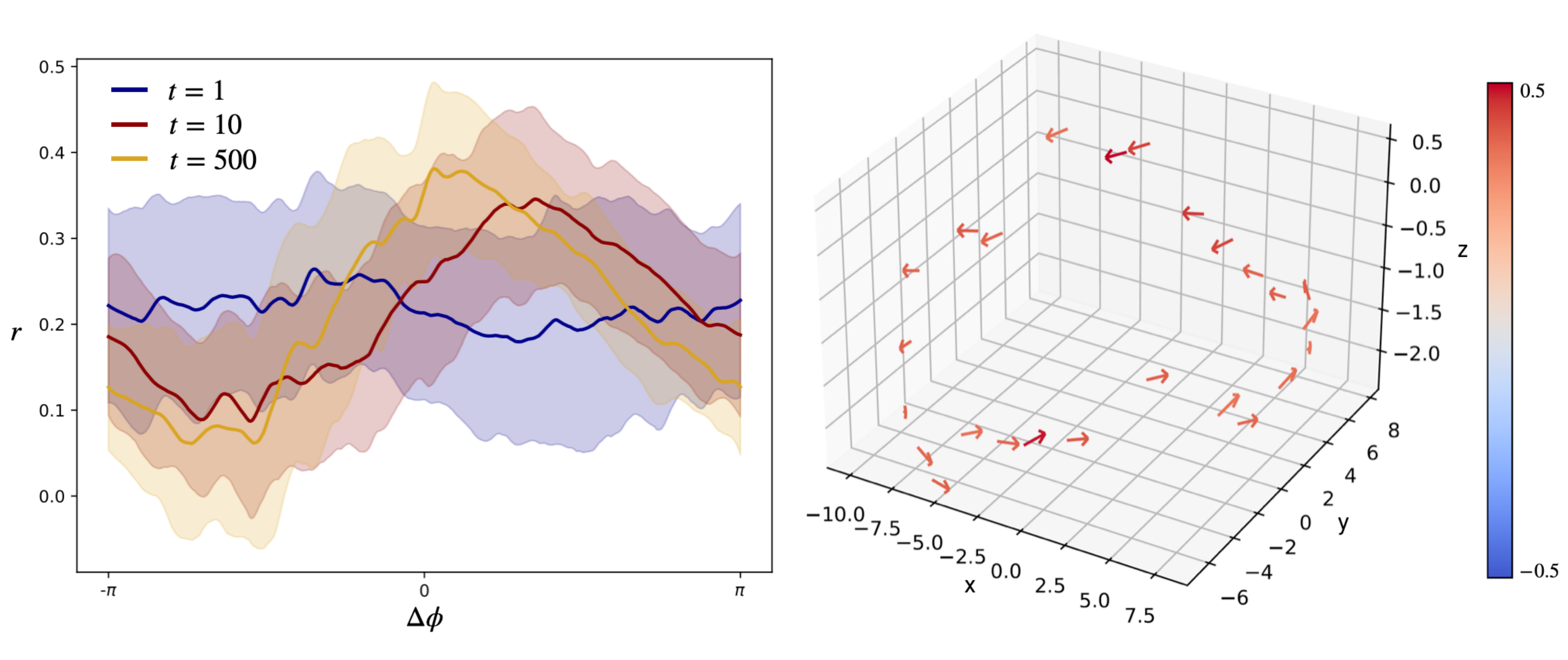}
    \caption{Evaluation of the reward function. Left: Sensitivity with respect to changes in the swimming direction $\Delta \phi\in[-\pi,+\pi]$ at different times $t\in\{1,10,500\}$. The bold lines indicate the reward averaged over all swimmers, and the shading shows $\pm \sigma$ confidence bounds on an example trajectory. Right: Individual rewards of the swimmers at $t=500$ in their original configuration.}
    \label{fig:reward-vis}
\end{figure}
As input to IMARL, we create a set of observed trajectories for milling. In order to ensure that our demonstration data has a high number of desired demonstrations, we maximize the time-averaged rotation by deploying CMA-ES~\cite{cmaes} on the three radii of the model. Using the resulting parameters, we perform simulations using 25 swimmers and 1000 steps per trajectory. We select 50 trajectories from the simulations with a time-averaged rotation larger than 0.80. \Cref{fig:demo-traj} shows two typical trajectories of the swimmers in the demonstration data $\mathcal{D}$. 

After collecting the demonstration dataset, IMARL is used to learn the reward function for the agents using five million interactions with the environment. The state and action are defined as follows: The state $\st^{(i)} \in \mathbb{R}^{35}$ of a swimmer consists of the distances and angles to the nearest neighbors and the angles between the swimming directions. We set the number of nearest neighbors to 7. The action $\ac^{(i)}\in\mathbb{R}^2$ is the update of the swimming direction. The maximal change in direction is $\pm 4^\circ$, the velocity of the particles is constant $v=\|v_i\|=3.0$, and time is discretized into intervals of length $\Delta t=0.1$, analogously to \cite{Gautrais2008}. We plot the time-averaged rotation for each trajectory during training in \cref{fig:training}. After approximately one million steps, the averaged rotation reaches the asymptotic value of 0.8. On the right, we plot the return of the agents calculated from the learned reward function $\re_\vartheta$ for the trajectories in the demonstration and background batches. We see that the demonstrations in $\mathcal{D}$ reach the maximal return of 500, whereas the trajectories in the background batch have a lower return. In \cref{fig:irl-traj}, we show two trajectories sampled from the learned policy $\pi_\omega$ that exhibit a time-averaged rotation larger than 0.80. The trajectories capture the behaviour in the demonstration data: milling with a sideways drift (upper) and milling at a fixed location (lower). To analyze the reward function $\re_\vartheta$, we change the swimming direction by performing a rotation $\Delta \phi$ around the z-axis to each swimmer individually and we calculate the reward for the altered states. We plot the rewards' mean and standard deviation in~\cref{fig:reward-vis}). We observe that the sensitivity of the rewards at the undisturbed configuration ($\Delta \phi = 0$) increases with time. During initialization ($t=1$), when the configuration is randomized, the reward is insensitive to changes in the state. When the swarm reaches a milling configuration ($t=500$), the sensitivity and the reward are highest. Therefore we conclude that the swimmers converge to their preferred configuration. To the right of \cref{fig:reward-vis}, we plot the reward of the swimmers at $t=500$ in their original configuration. We observe an almost uniform distribution of rewards among the swimmers, which can be explained by the fact that the trajectory is close to the demonstrations. Therefore, the return is almost maximal for each swimmer.

\section{Conclusion}
\label{sec:conclusion}

Understanding and quantification of the collective behavior in complex systems such as crowds~\cite{Helbing2005}, markets~\cite{Vriend1995}, animal herds and fish schools~\cite{Aoki1982}, and even artificial multi-agent systems~\cite{multiagentsystems} has been a long standing challenge across several disciplines. Advances in machine learning offer new perspectives and tools that can  address these challenges. In this work we showcase this potential by focusing on particle models of fish schooling behaviour. Prior work has shown that the observed swarming, schooling, and milling patterns can be distinguished based on order parameters such as rotation and polarization. Although carefully developed particle models can be tuned to allow recovering these patterns~\cite{Aoki1982,Gautrais2008}, an automated method is desirable. The requirement for such an automated discovery is that the resulting agents act solely based on local information. 

Here we automate  the discovery of local incentives based solely on observed trajectories, through a novel combination of MARL with GCL. To the best of our knowledge this is the first off-policy IMARL algorithm for a continuous state and action space. We demonstrate that the present method can recover the performance of the forward problem while automatically discovering a suitable reward function approximated through a neural network using OpenAI gym MuJoCo environments as a benchmark. We then showed the applicability of discovering local rewards from synthetic demonstrations of collective behavior. We find that our method can successfully recover the behavior present in the data. This indicates it's potential to a broad range of systems with collective behavior. While the learned neural network offers limited interpretability, we demonstrate how we can examine it's characteristics when systematically varying the state of the individuals.

We conclude this  discussion by addressing some limitations of IMARL. Even though IRL removes the burden of defining the reward function, one of the remaining challenges lies in determining the state and actions, which are typically required to be specified by the user. Additionally, selecting hyperparameters in IMARL requires careful consideration, as the learning success and convergence speed are highly sensitive to these choices. Furthermore, IMARL involves significant computational expenses, particularly in terms of memory requirements, when compared to traditional RL methods. 
Lastly, although employing a neural network for the reward function offers great flexibility, it comes at the cost of reduced interpretability in the obtained results. However, our aim is to empirically demonstrate the general claim that the reward function is the most transferable object in a decision-making process.

\begin{ack}
We acknowledge the infrastructure and support of CSCS, providing the necessary computational resources under project s1160.
\end{ack}

\bibliographystyle{plain}
\bibliography{bibliography.bib}








\end{document}

%% file: macros.tex
\newcommand{\st}{s}
\newcommand{\stS}{{\cal S}}

\newcommand{\ac}{a}
\newcommand{\acS}{{\cal A}}

\newcommand{\re}{r}

\newcommand{\sti}{s^{(i)}}

\newcommand{\vstate}{\boldsymbol{\st}}
\newcommand{\vaction}{\boldsymbol{\ac}}
\newcommand{\vreward}{\boldsymbol{\re}}

\newcommand{\DATA}{\mathcal{D}}
\newcommand{\BCK}{\mathcal{B}}
\newcommand{\RF}{\mathcal{R}_\vartheta}

\newcommand{\COMMA}{\,,}
\newcommand{\PERIOD}{\,.}
\newcommand{\GIVEN}{\,|}

\newcommand{\loc}{\boldsymbol{x}}
\newcommand{\vel}{\boldsymbol{v}}
\newcommand{\disp}{\boldsymbol{r}}